\documentclass[letterpaper, 10 pt, conference]{ieeeconf}  

\IEEEoverridecommandlockouts                              

\usepackage{amsmath}
\usepackage{amssymb}  
\usepackage{booktabs}
\usepackage{censor}
\usepackage{float} 
\usepackage{epsfig} 
\usepackage{algpseudocode}
\usepackage{siunitx} 
\usepackage[dvipsnames]{xcolor}
\usepackage{algorithm}
\usepackage{hyperref}
\hypersetup{
    colorlinks=true,
    linkcolor=blue,
    urlcolor=blue,
    citecolor=blue,
    pdftitle={ToolCo-DesignAutoChemLab},
    pdfpagemode=FullScreen,
    }
\usepackage{nicematrix}
\usepackage{colortbl}
\newcolumntype{C}{>{\centering\arraybackslash}X}
\usepackage{gensymb}
\usepackage{multirow, array}
\usepackage{caption}
\usepackage{subcaption}
\usepackage[utf8]{inputenc}
\makeatletter
\long\def\@makefntext#1{\parindent 0pt\noindent#1}
\makeatother

\usepackage{microtype}
\title{\LARGE \bf Tool-Policy Co-Design for Powder Weighing in Laboratory Automation}

\author{Nikola Radulov$^{1}$, Xin Yang$^{1}$, Kevin S. Luck$^{3}$ and Gabriella Pizzuto$^{1, 2}$
\vspace{-0.5em}
\thanks{{$^{1}$ School of Computer Science \& Informatics, University of Liverpool, UK.}}
\thanks{$^{2}$ Department of Chemistry, University of Liverpool, UK.} 
\thanks{$^{3}$ Vrije Universiteit Amsterdam, Faculty of Science, Netherlands.}
\thanks{This work was supported by the Leverhulme Trust through the Leverhulme Research Centre for Functional Materials Design, the Royal Academy of Engineering under the Research Fellowship Scheme and EPSRC through the New Investigator Award (UKRI2999).}}

\begin{document}
\maketitle

\begin{abstract}
Autonomous powder weighing is one of many bottlenecks in laboratory automation due to the complex, non-linear dynamics of heterogeneous materials. 
Robot chemists performing this task utilise standard tools shaped for the dexterity of human hands, whose fixed geometry sets the dynamics that the control policy needs to regulate.
This work introduces a tool-policy co-design framework that concurrently optimises the morphology of a dispensing tool and its control policy for use by robots in chemistry laboratories, formulated as a bi-level optimisation that minimises dispensing error over a target distribution of powder flowabilities. 
The outer loop varies tool-design parameters such as tool depth, width and rim spike topology using Bayesian optimisation and hyperband, while an inner loop optimises a control policy for each candidate morphology. 
We also introduce a geometric similarity metric that warm-starts policy training from cached policies of structurally similar designs, exploring $28\%$ more configurations under the same compute budget. 
The proposed framework is evaluated on a robotic powder weighing task across seven materials with distinct physical dynamics in a flowability-informed robot-material simulation framework.
Experimental results demonstrate that our co-designed tool morphology reduces real-world weighing errors by $45\%$ relative to a standard tool, including on previously unseen materials.  
These results demonstrate our method can adapt both the control policy and the physical tool to the dynamics of the target material, bringing a new paradigm for material manipulation to the field of laboratory automation.

\end{abstract}
\section{Introduction} 

Accelerating the discovery of materials is crucial for addressing global challenges and self-driving laboratories (SDLs) pursue this by integrating automated hardware with data-driven decision making~\cite{sdl_review}. 
To date, laboratory robots have been deployed predominantly for sample transportation, where the manipulated object is rigid.
Sample preparation remains a fundamental challenge, as materials deform and flow under contact and their response is governed by bulk properties that differ across materials.
Robot-material manipulation underpins solid-state chemistry~\cite{sdl_review}, yet robots inherit tools such as spatulas~\cite{omron_powder_weighing, radulov2026flipflowabilityinformedpowderweighing} shaped for use by the dexterous human hand rather than for a manipulator or the material.
Recent approaches use deep reinforcement learning (RL) to adapt robot manipulators to these dynamics~\cite{ omron_powder_weighing, radulov2026flipflowabilityinformedpowderweighing}, which optimise a control policy around a fixed tool. 
Tool geometry defines the action-to-mass transfer function, bounding the task precision achievable by any control policy using that tool.
While co-design of morphology and control is increasingly explored, it remains largely unaddressed for tool-material manipulation which is fundamental to the success of robot-driven chemistry lab automation.

\begin{figure}[]
    \centering
    \includegraphics[width=0.4\textwidth]{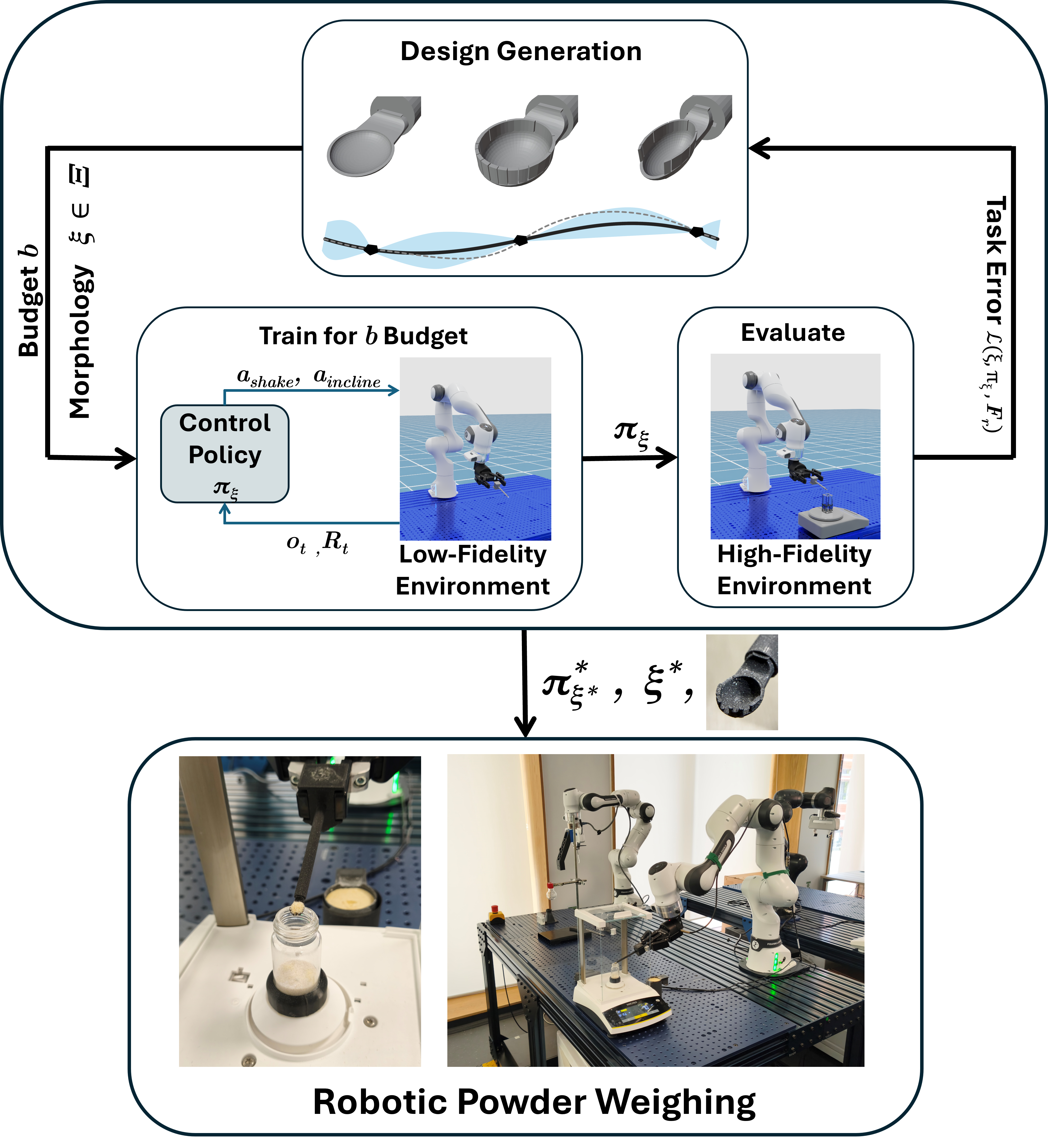}
    \caption{Tool-policy co-design method for powder weighing. We use BOHB to propose tool morphologies $\xi$. For each candidate $\xi$, a control policy $\pi_\xi$ is trained in a low-fidelity environment and evaluated in a high-fidelity digital twin. The highest-performing design and paired policy are transferred for real-world validation.}
    \label{fig:overall_blockdiagram}
\end{figure}

To address this, we present a co-design framework that concurrently optimises both the morphology of a dispensing tool and the robot's reinforcement learning control policy, formulated as a bi-level optimisation problem that minimises dispensing error across a target distribution of powder flowabilities (Fig.~\ref{fig:overall_blockdiagram}).
In an outer loop, Bayesian optimisation and hyperband (BOHB) explores a parameterised, low-dimensional morphological search space defining the depth, width, and discrete rim spike topology of an ellipsoidal tool. 
We select BOHB over standard Bayesian optimisation because successive halving evaluates candidates at low training budgets and promotes only top performers, saving compute on sub-optimal morphologies.
In the inner loop, an RL control policy is trained for each candidate tool. 
To make the search tractable, we introduce a geometric similarity metric that warm-starts policy training from cached policies of structurally similar designs.
We validate the framework in simulation and by zero-shot transfer to a physical robot.

In summary, the main contributions of this work are:
\begin{itemize}
    \item A bi-level co-design framework for milligram-scale powder weighing that jointly optimises a reinforcement learning control policy and the tool morphology across a distribution of material flowabilities, over a parameterised design space derived from standard tool geometries that combines continuous bowl scaling with discrete rim spike configurations.
    \item A morphological similarity metric that accelerates the search by warm-starting policy training from structurally related candidates, without compromising the real-world performance of the discovered tool.
    \item An empirical evaluation in simulation and on a real robotic manipulator, demonstrating that the co-designed tools outperform both a standard tool and grid-search baselines across a diverse range of powder behaviours.
\end{itemize}

\section{Related Work}
\subsection{Robot Skill Learning for Laboratory Automation}
Robotic scientists have transitioned from open-loop automation to adaptive modular systems, motivated by the need for generalisable approaches capable of handling unpredictable, heterogeneous samples~\cite{sdl_review}.
While self-driving labs have advanced high-level planning and experimental decision making~\cite{sdl_review}, low-level manipulation remains a fundamental bottleneck.
Sample grinding~\cite{grinding}, powder weighing~\cite{omron_powder_weighing} and scooping~\cite{adaptive_scooping} are challenging due to the non-linear, unpredictable dynamics of materials.
To address this, recent works leverage deep reinforcement learning through sim-to-real domain randomisation~\cite{omron_powder_weighing} or embed material properties directly into the learning process~\cite{radulov2026flipflowabilityinformedpowderweighing}.
Prior work optimises a control policy around a fixed, human-centric tool, so task success is bounded by a geometry that was not designed for robot-material manipulation.
We instead formulate the tool as a design variable of the robot skill itself.

\subsection{Robotic Tool Co-Design}
Co-design jointly optimises a robot's morphology and its control policy, on the premise that a fixed embodiment bounds control performance.  
Recent works search a parametrised design space, either through bi-level formulations pairing Bayesian optimisation with reinforcement learning, \textit{e.g.} to design mobile manipulator mountings~\cite{schneider_co_design}, or through generative models and cross-embodiment policies combined with grammar-based search to scale the synthesis of tool shapes~\cite{lin2025robotsmithgenerativerobotictool}.
Closest to our work, Li \textit{et al.}~\cite{Li2023} optimise tool geometry for contact-rich tasks over a distribution of task variations using differentiable simulation. 
However, they vary the initial object pose instead of its dynamics and hold the policy fixed, leaving joint optimisation under such variation unaddressed. 
Within laboratory automation, specialised end-effectors have been engineered \textit{e.g.}, the SCU-Hand uses a soft, reconfigurable conical sheet for adaptive scooping~\cite{11128802}, later extended with a single-sheet valve for milligram-scale dispensing~\cite{takahashi2026scuhandintegratedsinglesheetvalve}. 
However, the morphologies of these tools are manually handcrafted. 
Our approach concurrently optimises the control policy alongside the tool's physical design, tailoring the tool to the flow properties of the material.

\section{Methodology}
\label{sec:methodology}

Our method formulates robotic powder weighing as a joint optimisation problem with bi-level structure over tool morphology and control policy. 
The goal is to identify a tool design that minimises the expected dispensing error over a distribution of materials with diverse physical dynamics.

\subsection{Problem Formulation}
\label{ssec:problem_formulation}

We consider the task of autonomous powder weighing with a robotic manipulator that holds a custom tool at its end-effector and dispenses powder into a container on an analytical balance (Fig.~\ref{fig:overall_blockdiagram}). 
The balance provides weight feedback at each control step and an episode terminates after a fixed number $T$ of steps. 
Following Radulov \textit{et al}.'s method~\cite{radulov2026flipflowabilityinformedpowderweighing}, we characterise a material by its static flowability. 
This is quantified by the angle of repose (AoR), where a larger AoR corresponds to a lower flowability.
The flowability $F$ of a material is sampled from a representative range $F_r=[AoR_{min}, AoR_{max}]$ across the task distribution.

Let the morphology of the tool be described by a parameter vector $\xi \in \Xi$, where $\Xi$ denotes the design space defined in Section~\ref{ssec:tool_generation}.
Let $\pi_\xi$ be the robot control policy trained with that tool.
As the geometry of the tool determines how much powder is displaced per action, $\xi$ and $\pi_\xi$ cannot be chosen independently, \textit{i.e.}, a morphology is only as good as the policy that can be learnt for it and vice versa.
Furthermore, powder dispensing is difficult to mathematically model as the dynamics vary strongly with material properties and with the amount of powder remaining in the tool~\cite{omron_powder_weighing, radulov2026flipflowabilityinformedpowderweighing}.

To find the optimal combination $(\xi^*, \pi^*_{\xi^*})$ we define the bi-level optimisation problem as:
\begin{align}
    \xi^* = \arg \min_\xi \mathbb{E}_{F_r}[\mathcal{L}(\xi, \pi^*_\xi, F_r) ], \\
    \text{s.t.~} \pi^*_\xi = \arg \max_\pi \mathbb{E}_{\substack{
        \pi(a_t \vert s_t) \\
        F\sim F_r \\
        p(s_{t+1} \vert s_t, a_t, \xi, F)
        }}\left[G \right]
\end{align}
Here, we optimise the morphology $\xi$ and policy $\pi_\xi$ using a morphology objective $\mathcal{L}$ and a policy learning objective $G$.
The complete bi-level co-design framework is illustrated in Fig.~\ref{fig:overall_blockdiagram}.






\subsection{Outer Loop: Tool Morphology Parametrisation} 
\label{ssec:tool_generation}

\begin{figure}[h!]
    \centering
    \includegraphics[width=0.6\linewidth]{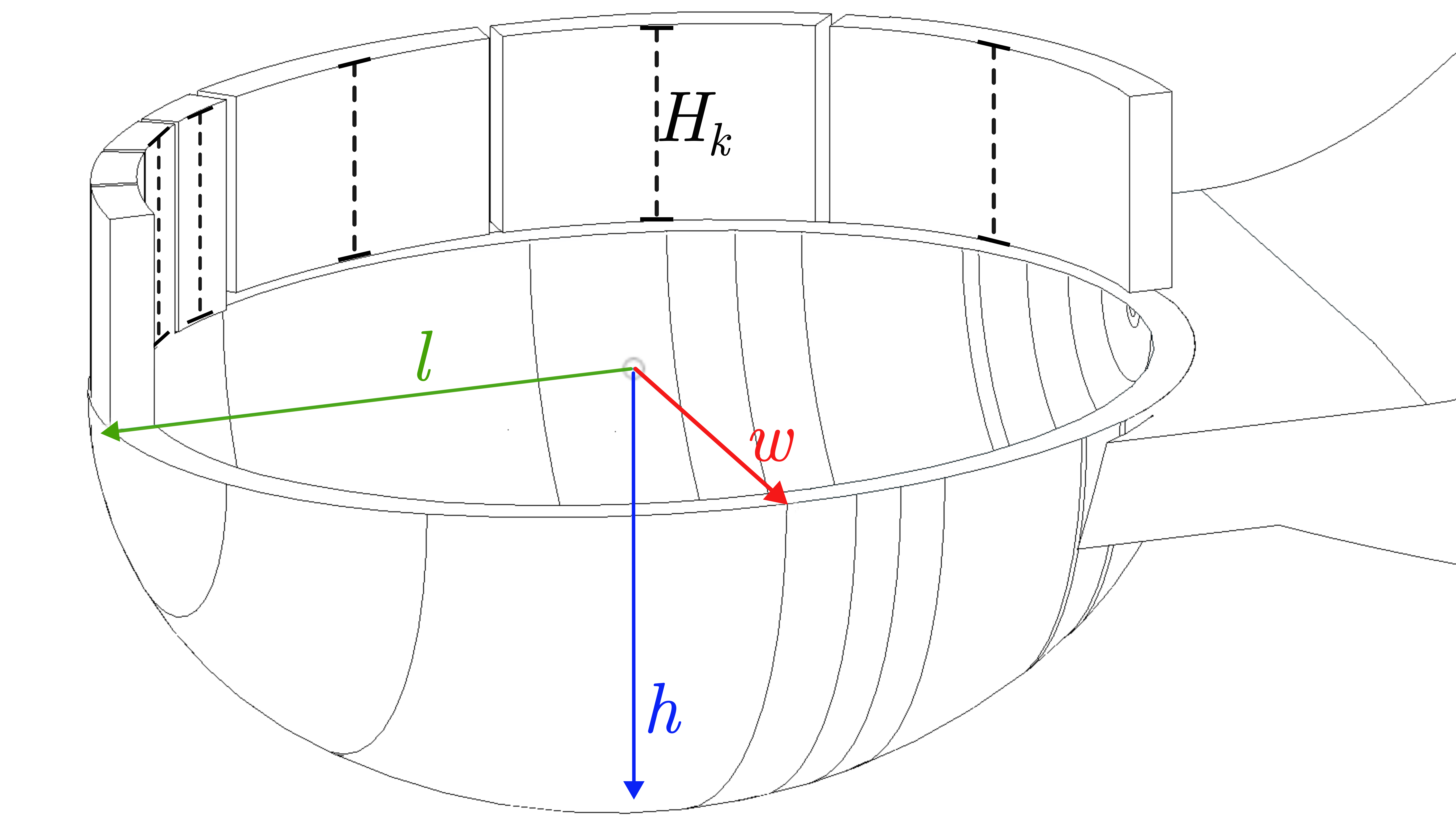}
    \caption{Tool morphology parameters of the design space; $l$, $h$, $w$ represent the length, depth and width of the spoon bowl. $H_k$ represents the height of the $k^{th}$ spike.}
    \label{fig:design parameters}
\end{figure}

We base the design space on a standard spoon-like geometry, where the bowl is modelled as an ellipsoid of height $h$, width $w$, and length $l$, sliced at the equator. 
Three groups of parameters, illustrated in Fig.~\ref{fig:design parameters}, form $\xi$.

\begin{enumerate}
    \item \textit{Tool depth ($h$)} governs the vertical extent of the spoon bowl. A shallower depth yields a flatter geometry, requiring less aggressive agitation and smaller pitch adjustments, which can be ideal for cohesive materials.
    
    \item \textit{Tool width ($w$)} controls the lateral scale of the bowl. Increased widths broaden the surface area, enabling the capture of larger volumes per scoop.
    
    \item \textit{Tool rim spike heights ($H_k$)} are used for finer control over material outflow. The outer optimisation loop can independently vary the height of each spike. As the spikes are modelled along the edge of the bowl, their spatial distribution is inherently coupled with the \textit{tool depth}.
\end{enumerate}

The bowl length $l$ is constant, which keeps the tool centre point identical for all candidates.
Should $l$ be allowed to vary, the parameters of the low-level Cartesian position controller executing the shake and incline primitives would have to be retuned per candidate, in simulation and on the robot, to avoid collisions with the vial and analytical balance.
Since the bowl volume varies across $\Xi$, the granular simulation is configured so that its fidelity is equivalent for every candidate.

\subsection{Outer Loop: Tool Design Generation} 
\label{ssec:outer loop}
We search the morphological space $\Xi$ with BOHB~\cite{Falkner2018BOHB}.
This method proposes candidates from a surrogate model fitted to past evaluations and allocates a budget $b$, defined as the number of training episodes granted to the inner loop.
Budgets are assigned by successive halving, where a large pool of morphologies is trained on a small budget, the worst performers are discarded and the survivors are promoted to a longer budget. 
At each outer-loop iteration, BOHB samples a candidate tool design $\xi \in \Xi$. 
This configuration is passed to the robotic simulator, which applies the required scaling to instantiate the custom tool geometry before initiating the inner loop to train a corresponding control policy for $b$ episodes. 
Due to the stochasticity inherent in policy optimisation, we evaluate $n$ random seeds per configuration.
We define the outer-loop objective function $\mathcal{L}(\xi,\pi^*_\xi, F_r)$ as the expected weighing error across a flowability range $F_r$. 

 \begin{equation}
    \mathcal{L}(\xi, \pi^*_\xi, F_r) =  \mathbb{E}_{F \in F_r} \left| w_{\mathrm{target}} - w_{\pi^*_{\xi}, T}^{F} \right|
    \label{eq:scoring_func}
\end{equation}

where $w_{\mathrm{target}}$ represents the target weight to be dispensed, and $w_{\pi^*_{\xi}, T}^F$ represents the final dispensed weight for material $F$, using the candidate tool design $\xi$ and its associated control policy $\pi^*_{\xi}$.
In practice, to increase stability we sample over a distribution/set of optimised policies (\textit{i.e.}, multiple seeds), \textit{i.e.}, $\mathcal{L}_\text{avg} = \frac{1}{n}\sum_{\pi^*_\xi \in \Pi^*_\xi} \mathcal{L}(\xi, \pi^*_\xi, F_r)$.

\subsection{Inner Loop: Policy Optimisation}
\label{ssec:policy_optimisation}
Given a morphological design $\xi$, the aim of the inner loop is to optimise a corresponding control policy $\pi_\xi$ and provide a reliable estimate of its performance over the target flowability range $F_r$. 
We formulate the powder weighing task as a Markov decision process (MDP)~\cite{sutton2018} $\mathcal{M} = \langle \mathcal{S}, \mathcal{A}, \mathcal{P}, \mathcal{R}, \gamma \rangle$.
We employ the same parameterised action space as prior works~\cite{omron_powder_weighing, radulov2026flipflowabilityinformedpowderweighing}, where at each step, the agent adjusts the tool pitch by $a_{\mathrm{incline}}$, and subsequently performs a shaking motion by retracting the tool a distance $a_{\mathrm{shake}}$ before returning it to its home position. 
The observation $o \in \mathcal{S}$ follows prior work: $o=(w_{current},w_{target}, \theta_{spoon})$.
We introduce two enhancements from Radulov \textit{et al}.'s work~\cite{radulov2026flipflowabilityinformedpowderweighing}. 
\subsubsection{Reward}
We redesign the reward to penalise the step-wise change in error rather than the absolute error at the current step:
\begin{equation} 
    \mathcal{R}_t = \frac{\Delta_{t-1} - \Delta_{t}}{w_{\mathrm{target}}}
    \label{eq:reward}
\end{equation}
where $\Delta_t = |w_{\mathrm{target}} - w_{t}|$ is the absolute weight error at step $t$, with the initial condition $\Delta_0 = w_{\mathrm{target}}$. 
This formulation telescopes such that the cumulative return over an episode becomes $G = \sum_{t=1}^T \mathcal{R}_t = 1 - \frac{\Delta_T}{w_{\mathrm{target}}}$, which depends only on the final dispensing error.
Under the original step-wise absolute error reward, the inner loop implicitly optimised for speed as well as accuracy, encouraging the agent to close the mass gap rapidly and sometimes overshoot.
Since our outer loop scores a morphology strictly on final precision $\mathcal{L}$ (Equation~\ref{eq:scoring_func}), this reward corrects the mismatch between the inner loop's objective and the outer loop's cost function.


\subsubsection{Sampling of the Flowability Range}
We replace the curriculum learning strategy with uniform random sampling of the flowability range $F_r$. 
A curriculum makes the training distribution a function of the budget $b$ which conflicts with the successive halving of BOHB in our outer loop.
At low budgets the agent would only be exposed to the easy end of $F_r$, so Equation~\eqref{eq:scoring_func} would rank morphologies on a narrow subset of materials, shifting the ranking as candidates are promoted.
Uniform sampling makes the score at every budget an unbiased estimate of the same quantity, so that evaluations obtained at different fidelities remain comparable.


\subsection{Similarity-based Evaluation} 
\label{ssec: similarity}

Evaluating the outer loop objective function $\mathcal{L}(\xi,\pi^*_\xi, F_r)$ presents a computational bottleneck as it requires training the control policy $\pi_\xi$ from scratch for every proposed tool morphology $\xi$. 
However, during the BOHB optimisation process, the algorithm frequently samples configurations that represent minor parametric perturbations of previously evaluated designs. 
As the physical dynamics of powder flow remain largely consistent across minor morphological adjustments, we leverage the fully trained policy of a structurally similar design to warm-start the training of the new control policy. 
To facilitate this accelerated search, we introduce a morphological similarity metric $S(\xi_1, \xi_2) \in [0,1]$ that quantifies the resemblance between two geometries based on their total volume, geometry and spike topology.

For both the total volume $V$ and the width-to-depth ratio $R$, we employ a min--max ratio to ensure scale-invariant similarity bounds within $[0,1]$. 
The total volume $V$ of a given morphology is computed as the sum of the ellipsoidal half-bowl volume and the volume of the watertight elliptical cylinder formed by the rim spikes. 
The height of this cylinder is bounded by the shortest spike.
The volume and ratio similarity scores are computed as:
\begin{equation}
    S_{V} = \frac{\min(V_1, V_2)}{\max(V_1, V_2)}, \quad S_{R} = \frac{\min(R_1, R_2)}{\max(R_1, R_2)}
    \label{eq:sim_vr}
\end{equation}

With spike height already captured volumetrically, the final term measures perimeter topology, where we weight the squared differences between the first-order discrete derivatives of adjacent spikes by their arc ratio $c_i$ and map the result into $[0, 1]$ with a negative exponential function:
\begin{equation}
    S_{\mathrm{spikes}} = \exp \left( - \sum_{i=1}^{M} c_i (\delta_{1,i} - \delta_{2,i})^2 \right)
\end{equation}

The overall similarity score $S(\xi_1, \xi_2)$ is defined as a weighted linear combination of $S_V$, $S_R$, and $S_{\mathrm{spikes}}$:

\begin{equation}
    S(\xi_1, \xi _2) = \alpha S_{V} + \beta S_R + \gamma S_{\mathrm{spikes}}
    \label{eq:sim_total}
\end{equation}

Here, $\alpha + \beta + \gamma = 1$. 

We incorporate the similarity metric into our optimisation method as detailed in Algorithm~\ref{alg:bohb_spoon}, where the similarity logic is implemented in lines 10 to 16. 
Throughout the optimisation, the system maintains an observation history $\mathcal{H}$ to train the Bayesian optimisation surrogate model and a policy registry $\mathcal{P}$ to cache learned control policies. 
The algorithm first calculates the maximum bracket index $s_{\max}$, which dictates the maximum number of successive halving stages. 
BOHB iterates through several brackets, starting from the most aggressive one with $s_{\max}$ successive halvings.
At the start of a stage, a set of $N$ initial morphologies $\mathcal{C}$ is sampled from a Bayesian surrogate model $\mu(\xi|\mathcal{H})$ conditioned on prior history. 
Within each successive halving stage $k$, the target budget $b$ is computed. 
Before training a policy $\pi_{\xi_i}$ for a morphology $\xi_i$, the algorithm computes the similarity score $S(\xi_i, \xi_j)$ (line 10) against all previously trained morphologies $\xi_j \in \mathcal{H}$. 
It filters for a candidate set $\mathcal{S}$ of morphologies that meet a minimum similarity threshold $\sigma$ and precisely match the target budget $b$. 
If a valid match is found, the policy $\pi^* \in \mathcal{P}$ of the most similar morphology $\xi^*$ is used to warm-start $\pi_{\xi_i}$ and the execution budget $b$ is reduced to $\phi$. 
Conversely, if $\mathcal{S}$ is empty, the policy undergoes standard training for the full budget $b$. 
After training, the policy is evaluated to determine its task error $e_i$ and both the registry $\mathcal{P}$ and history $\mathcal{H}$ are updated. 
Finally, the configuration pool $\mathcal{C}$ is pruned, retaining only the top $N_k$ performers to advance to the next iteration.

\begin{algorithm}[]
\caption{Tool-Policy Optimisation for Powder Weighing}
\label{alg:bohb_spoon}
\begin{algorithmic}[1]
\Require morphology search space $\Xi$, maximum budget $b_{max}$, minimum budget $b_{min}$,
 similarity threshold $\sigma$, warmup fraction $\phi$, successive halving proportion $\eta$, flowability range $F_r$.

\State $\mathcal{H} \gets \emptyset$
\State $\mathcal{P} \gets \emptyset$
\State $s_{max} \gets \lfloor \log_\eta\left(\frac{b_{max}}{b_{min}}\right) \rfloor$
\For{$s = s_{max}, s_{max}-1 \dots 0$}
    \State $N \gets \lceil \frac{s_{max}+1}{s+1} \cdot \eta^s \rceil$ 
    \State $\mathcal{C} \gets \{\xi_i\}_{i=1}^{N} \sim \mu(\xi|\mathcal{H})$ 
    \For{$k = 0, \dots, s$} 
        \State $b \gets b_{max} \cdot \eta^{k-s}$ 
        \For{$\xi_i \in \mathcal{C}$}
            \State $\mathcal{S} \gets \{ ( \xi_j) \in \mathcal{H} \mid S(\xi_i, \xi_j) \geq \sigma $
            \Statex $\hfill \land \text{budget}(\pi_{\xi_j}) =b \}$
            \If{$\mathcal{S} \neq \emptyset$}
                \State $\xi^* \gets \arg\max_{\xi_j \in \mathcal{S}} S(\xi_i, \xi_j)$ 
                \State $\pi^* \gets \mathcal{P}[\xi^*]$
                \State $\pi_{\xi_i} \gets \pi^*$ 
                \State $b \gets  b \cdot \phi$ 
            \EndIf
            \State $\pi_{\xi_i} \gets \text{Train}(\pi_{\xi_i}, b)$
            \State $e_i \gets \mathcal{L}(\xi_i, \pi_{\xi_i}, F_r)$
            \State $\mathcal{P} \gets \mathcal{P} \cup \{\pi_{\xi_i}\}$ 
            \State $\mathcal{H} \gets \mathcal{H} \cup \{(\xi_i, b, e_i)\}$ 
        \EndFor
        \State $N_k \gets \lfloor N \cdot \eta^{-k} \rfloor$ 
        \State $\mathcal{C} \gets \text{TopK}(\mathcal{C}, N_k)$ 
    \EndFor
\EndFor

\end{algorithmic}
\end{algorithm}
\section{Experimental Evaluation}

In this section, we evaluated the proposed co-design framework in simulation and on a physical robot, with experiments designed to address the following research questions: (1) Can simulation predict real-world task performance and the relative ranking of tool morphologies? (2) Can BOHB co-design discover morphologies that outperform a standard (commercial) tool and grid-search baselines in the real world? (3) Does the morphological similarity threshold accelerate the search without compromising the tool performance?

\subsection{Experimental Setup}
\subsubsection{Simulation Setup}

Granular materials were simulated using the position-based dynamics (PBD) solver of NVIDIA Isaac Sim~\cite{NVIDIA_Isaac_Sim}, which resolves constraints such as contacts at the position level prioritising computational efficiency over physical fidelity and making it suitable for reinforcement learning. 
We constructed two environments (Fig.~\ref{fig:overall_blockdiagram}) to balance accuracy against cost.
The high-fidelity environment is a digital twin of our real-world setup and runs at a physics time step of $dt=1/240$\,\SI{}{s}. 
The low-fidelity environment removes the analytical balance and the target receptacle, reducing collision calculations per step and allowing a coarser time step of $dt=1/100$\SI{}{s}.
In both environments, a 3D model of the standard tool is attached to the robot's end-effector via a fixed joint matching the grasp pose of the real-world system. 
Episodes are initialised with the powder already contained within the tool, bypassing the computationally expensive tool-picking and scooping phases. 
To simulate how a tool's physical dimensions dictate its retained powder volume, we scale the baseline initial particle count bounds ($N_{\text{low}}, N_{\text{high}}$) proportionally with the volumetric capacity of the tool morphology $\xi$. 
Specifically, the initial particle count scales with the tool's width and its effective depth, accounting for both base bowl and the added volume from the spikes. 
Tool morphologies are instantiated by scaling the width and depth of the standard model.
The $M=14$ rim spikes are modelled as independent rigid bodies at fixed positions relative to the tool centre.
As dispensing is primarily directed through the tool tip, eight smaller spikes each spanning $c_i = 1/32$ of the ellipsoid arc are positioned there, while the remaining six span $c_i = 1/8$.
Spike height is set by vertical scaling to one of four discrete levels $[0, 0.33, 0.66, 1]$; if the morphological configuration $\xi$ sets a spike's height to zero, its corresponding rigid body is removed from the simulation.
The depth and width scale factors are continuous over $[0.7, 1.5]$ relative to the standard tool, so $\Xi$ combines two continuous scale factors with $4^{14}$ discrete spike configurations.
Episode length is fixed to $10$ steps. 
All experiments are distributed across four Ubuntu 22.04 workstations with NVIDIA RTX 4090/5090 GPUs; on the RTX 5090 machine, an episode averages \SI{13.57}{s} in the low-fidelity environment and \SI{24.95}{s} in the high-fidelity environment.

\subsubsection{Materials}
We defined the target material distribution over an AoR range $F_r=[28^\circ, 41^\circ]$, as in prior works~\cite{radulov2026flipflowabilityinformedpowderweighing}.
For our experiments, we sample materials at four fixed flowability points: $28^\circ$, $32^\circ$, $36^\circ$, and $41^\circ$. 
Highly cohesive materials (\textit{e.g.}, flour) are excluded from the range of simulated materials, as capturing their clumping dynamics would require alternative techniques that are prohibitively expensive for our framework.

\subsubsection{Real World Setup} 
\label{sssec: real life setup}

We use a Franka Research 3 (FR3)~\cite{franka} manipulator equipped with a Robotiq 85F gripper, mounted parallel to the working surface for precise tool manoeuvring. 
A Sartorius Entris II precision analytical balance measures the dispensed mass, feeding the data directly to the control policy in real time. 
The tool is loaded using a parabolic scooping trajectory~\cite{adaptive_scooping}, without the vision-based volume estimation, as calibrating it across all tool–material combinations is prohibitively expensive. 
Instead, we manually tune the scooping parameters so the initial acquired mass is within the simulated training distribution. 
To transfer a morphology from simulation, we fabricated the custom tool using a Prusa XL 3D printer equipped with a \SI{0.4}{mm} nozzle, which yields a vertical print resolution of \SI{0.1}{mm} and a horizontal resolution of \SI{0.4}{mm}. 
Consequently, the morphological search space $\Xi$ is discretised by these manufacturing constraints: the horizontal resolution sets the lower bound of $0.7$ on the scale factors, below which the rim spikes fall under the minimum reproducible feature size, while the vertical resolution sets the spacing of the four spike height levels.

\subsection{Sim-to-Real Transferability and Simulation Fidelity}
\label{ssec:grid_search}

To evaluate the capability of our simulation to predict real-world task performance, we perform a grid search across our morphological design space $\Xi$. 
We apply a step size of $0.4$ for both the depth and width scaling factors, which evaluates the extremities and the midpoint of these parameter ranges. 
We constrain the grid search to three predefined spike configurations: (1) all spikes set to maximum height; (2) all spikes disabled (height set to zero); and (3) a manually-designed configuration where the six front-most spikes are disabled while the remaining eight are set to their maximum height (represented by the array $\{0, 0, 0, 1, 1, 1, 1, 1, 1, 1, 1, 0, 0, 0\}$). 
To train the control policy $\pi_\xi$, we use Soft Actor-Critic (SAC)~\cite{Haarnoja2018}, a model-free, off-policy deep reinforcement learning algorithm. 
We adopt the neural network architecture and physics-informed modelling from prior work~\cite{radulov2026flipflowabilityinformedpowderweighing}, which includes seven optimised data points per flowability level. 
Training terminated after $3000$ episodes with $n=3$ random seeds per configuration. 
This is reduced from the $4000$ episodes in prior works, as standard tool policies converge well before the limit and the tighter budget also favours morphologies that remain sample efficient.

\begin{figure}[h!]
    \begin{subfigure}[a]{\columnwidth}
        \centering
        \includegraphics[width=\linewidth]{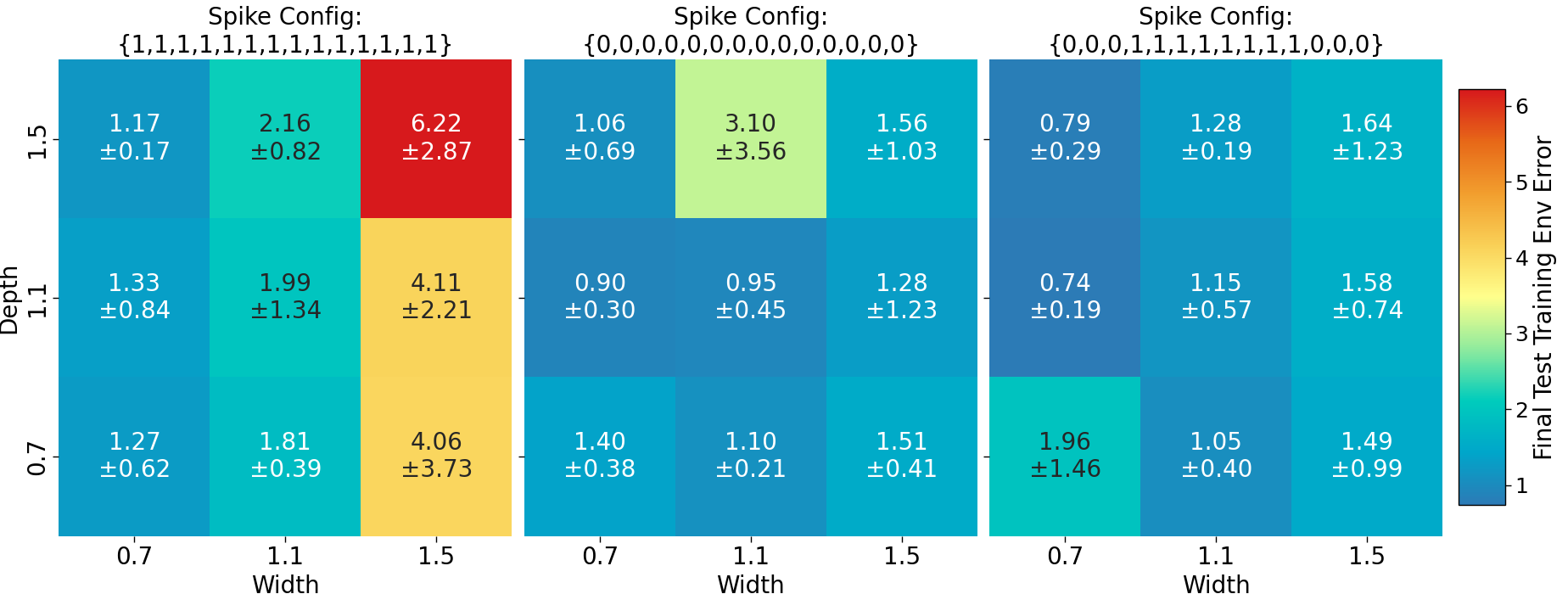}
        \caption{}
        \label{fig:error_fast_env}
    \end{subfigure}
    \begin{subfigure}[a]{\columnwidth}
        \centering
        \includegraphics[width=\linewidth]{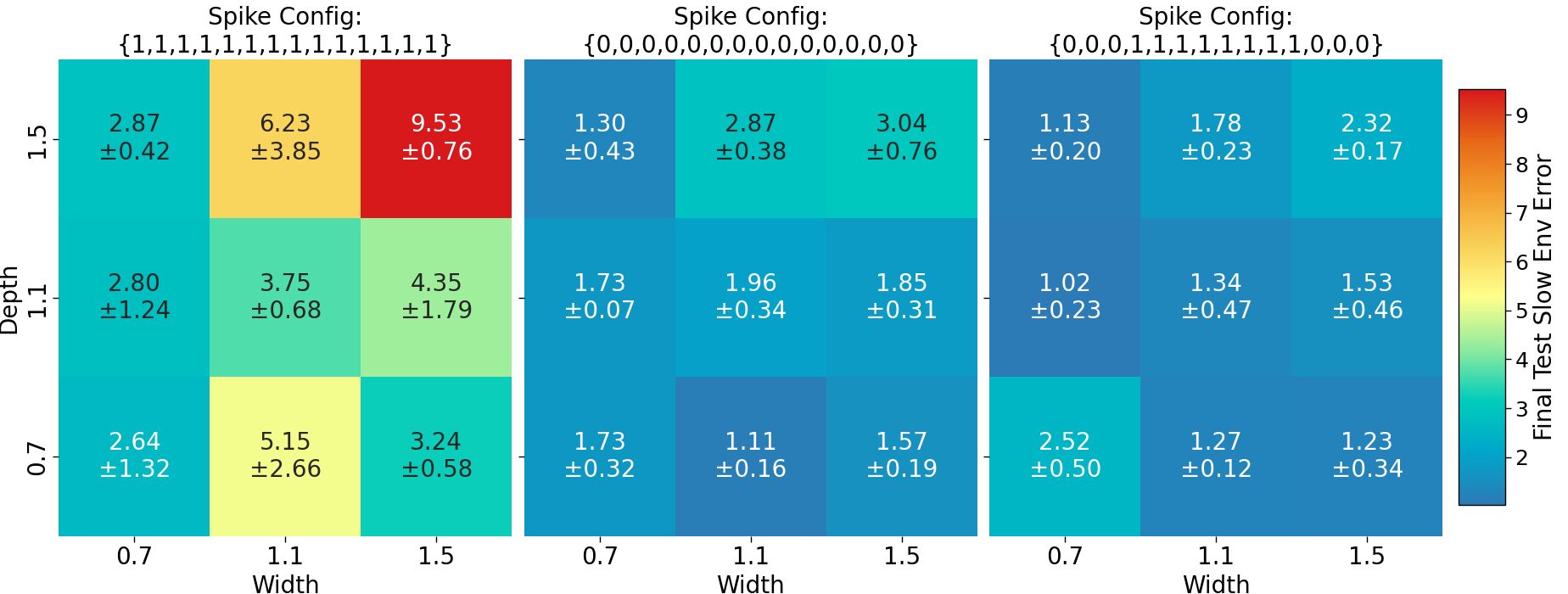}
        \caption{}
        \label{fig:error_slow_environment}
    \end{subfigure}
    \caption{The performance of each chosen morphology as the mean final test error and the respective standard deviation, averaged across 3 seeds.  The temperature represents the mean final test error (\SI{}{mg}). The top graph (\ref{fig:error_fast_env}) shows the raw task error when evaluated on the low-fidelity environment and the bottom graph (\ref{fig:error_slow_environment}) shows the performance when evaluated on the high-fidelity test environment.}
    \label{fig:grid_search_heatmap}
\end{figure}

For each configuration, we measured the empirical weighing error (Equation~\ref{eq:scoring_func}) across the sampled materials for a fixed target mass $w_{\mathrm{target}} = \SI{15}{mg}$, reporting the mean and standard deviation across seeds. 
While policy training occurred exclusively within the computationally-efficient low-fidelity environment, the final weight error was evaluated in both environments (Fig.~\ref{fig:grid_search_heatmap}). 
In both environments, tools with all spikes enabled performed worst, yielding a mean error of \SI{2.81}{mg} in the low-fidelity environment and \SI{4.69}{mg} in the high-fidelity environment. 
The manually-designed spike configuration achieves the lowest error in both cases: \SI{1.27}{mg} and \SI{1.54}{mg} respectively. 
In general, configurations with increased internal volume, \textit{i.e.}, those in which both depth and width are scaled up, performed worse than their smaller counterparts. 
This is likely because the larger retained powder mass makes it harder for the policy to exert fine-grained control over the dispensing flow.

To evaluate the simulation's predictive capability for tool optimisation, we selected nine morphologies for zero-shot transfer to the real setup, \textit{i.e.}, three from each spike configuration, encompassing the best- and worst-performing designs for the search. 
We evaluated the configurations on in-distribution (IID) materials and out-of-distribution (OOD) materials. 
We compare them against the standard tool, trained under identical conditions to those described in Section~\ref{ssec:policy_optimisation}.
The results are presented in Table~\ref{table:sim2real_results}. 
The most effective geometry is Config. \textbf{2}, which performs best on both the IID set and the full material set. 
Both Config. \textbf{2} and Config. \textbf{4} achieve IID errors lower than those of the standard tool. 
Config. \textbf{2} yields the lowest error for IID powders, apart from salt, where Config. \textbf{4} performs best. 

The sim-to-real prediction gap, defined as the absolute difference between simulated and physical error, averages $1.78 \pm 1.94$\SI{}{mg} for the high-fidelity environment and $2.63 \pm 2.83$\SI{}{mg} for the low-fidelity environment on IID materials. 
Over the entire test set (including OOD powders), this prediction gap widens to $4.18 \pm 1.78$\,mg and $5.10 \pm 2.34$\,mg respectively. 
This degradation is expected, as we explicitly exclude highly cohesive materials from simulation due to the prohibitive computational cost of modelling their complex dynamics, so the tool morphology cannot be optimised for clumping or compressible behaviour.

\begin{figure}[h!]
    \centering
    \includegraphics[width=0.9\linewidth]{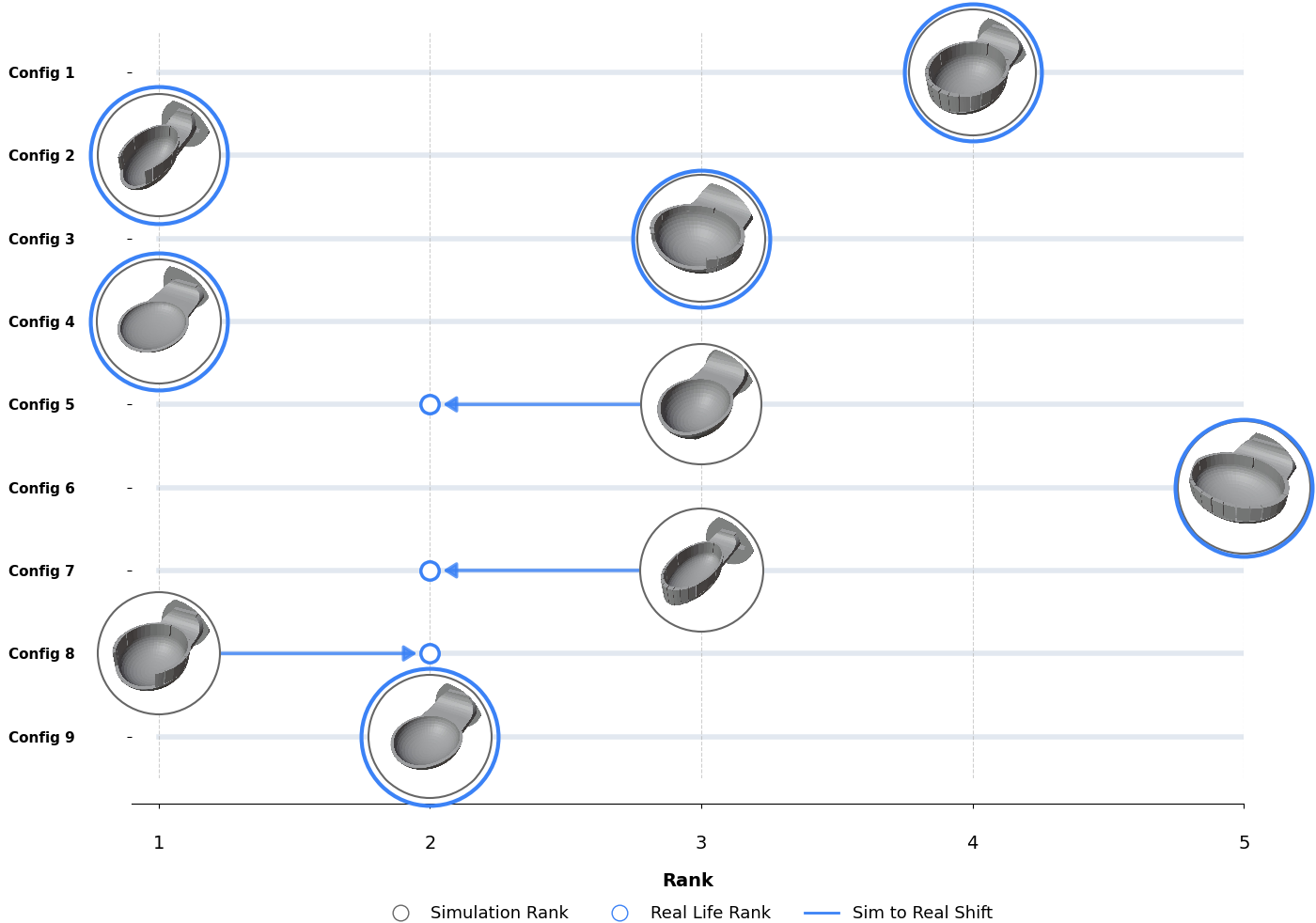}
    \caption{Sim-to-real rank shifts across morphologies. Each spoon image shows its baseline simulation rank; blue arrows show shifts for IID powders.}
    \label{fig:ranking_change}
\end{figure}

We ranked configurations by high-fidelity simulation performance using an anchor-based scheme: configurations are sorted by mean error, and each shares the current anchor's rank unless a one-tailed Student's t-test at 95\% confidence finds it significantly worse, where it becomes a new anchor. 
With $n=3$ seeds, configurations differing only slightly tend to share a rank.
Fig.~\ref{fig:ranking_change} illustrates the relative ranking shifts from simulation to the real world for IID materials. 
Only three of the tools change their standing: one simulation Rank 1 configuration drops to Rank 2, and two Rank 3 configurations are upgraded to Rank 2. 
The best and worst configurations by mean error retain their respective ranks across the sim-to-real gap, which supports the use of the simulation as a proxy for morphological optimisation.

\begin{table*}[htbp!]
    \caption{Zero-shot transfer results on $w_{target}=15$\SI{}{mg} of selected policy-tool pairs, evaluated using the best-performing policy in simulation for each morphology. For each powder, results report the mean absolute error and standard deviation over 10 real-world runs. Sodium bicarbonate, pectin and flour represent OOD materials, \textit{i.e.}, the policy was not trained on these specific powders or materials with similar flow. Data points marked with $*$ represent cases where no material was dispensed.}
    \centering
    \resizebox{0.99\linewidth}{!}{%
        \begin{NiceTabular}{l *{13}{c}}[hlines, cell-space-limits=3pt]
            \hline 
            \multicolumn{4}{c}{\Block{2-4}{\textbf{Spoon Configuration}}} & \multicolumn{7}{c}{\textbf{Powder Weighing Error (mg)}} & \Block{3-1}{In-distribution Error} & \Block{3-1}{Overall Error} \\
            & & & & \Block{2-1}{Sand} & \Block{2-1}{Sugar} & \Block{2-1}{Salt} & \Block{2-1}{Semolina} & \Block{2-1}{Sodium Bicarbonate} & \Block{2-1}{Pectin} & \Block{2-1}{Flour}\\
             No. & Depth & Width & Spike Configuration \\
            \hline 
              
            \midrule 
            \multicolumn{4}{c}{Standard Tool} & $1.49 \pm 1.36$ & $2.05 \pm 1.73$ & $1.66\pm1.19$ & $2.52\pm3.66$ & $4.45\pm5.57$ & $\mathbf{3.58\pm3.64}$ & $12.91\pm13.58$ & $1.93 \pm 2.16$ & $4.09\pm6.82$\\
            
            1 & 1.1 & 1.1 & \{1, 1, 1, 1, 1, 1, 1, 1, 1, 1, 1, 1, 1, 1\} & $2.12 \pm 4.21$ & $4.78 \pm 5.02$ & $12.56\pm2.98$ & $*$ & $14.48\pm0.23$ & $13.15\pm3.96$ & $13.29\pm4.06$ & $8.58 \pm 6.36$ & $10.74 \pm 5.79$ \\
            2 & 1.1 & 0.7 & \{0, 0, 0, 1, 1, 1, 1, 1, 1, 1, 1, 0, 0, 0\} & $\mathbf{0.67 \pm 0.63}$ & $\mathbf{0.53 \pm 0.53}$ & $1.01 \pm 0.54$ & $\mathbf{2.47 \pm 1.41}$ & $3.62 \pm 3.45$ & $8.21 \pm 4.04$ & $11.41\pm8.72$& $\mathbf{1.17 \pm 1.13}$ & $\mathbf{4.04 \pm 5.43}$\\
            3 & 1.5 & 1.5 & \{0, 0, 0, 1, 1, 1, 1, 1, 1, 1, 1, 0, 0, 0\} & $4.38 \pm 2.27$ & $2.29 \pm 1.27$ & $1.66\pm1.49$ & $14.48\pm 0.99$ & $14.49\pm0.97$ & $*$ & $14.66 \pm 0.34$ & $5.70 \pm 5.44$ & $9.52 \pm 6.06$ \\ 
            4 & 0.7 & 1.1 & \{0, 0, 0, 0, 0, 0, 0, 0, 0, 0, 0, 0, 0, 0\} & $0.81 \pm 0.38$ & $0.76 \pm 0.46$ & $\mathbf{0.56\pm0.40}$ & $2.76 \pm 2.57$ & $5.03\pm 4.08$ & $7.04 \pm 4.18$ & $13.70\pm1.4$ & $1.22\pm1.56$& $4.38\pm5.05$\\
            5 & 1.5 & 1.1 & \{0, 0, 0, 0, 0, 0, 0, 0, 0, 0, 0, 0, 0, 0\} & $0.90\pm0.53$ & $1.18\pm0.46$ & $2.23\pm1.20$ &$4.73\pm3.76$ & $9.16\pm8.89$ & $9.78\pm5.53$ & $15.71\pm14.68$ & $2.26\pm2.45$ & $6.24\pm8.42$\\
            6 & 1.5 & 1.5 & \{1, 1, 1, 1, 1, 1, 1, 1, 1, 1, 1, 1, 1, 1\} & $14.26\pm 2.14$ & $14.12\pm 1.87$ & $*$& $*$&$*$&$*$& $*$ & $14.59\pm 1.42 $ & $14.76\pm 1.09$\\
            7 & 0.7 & 0.7 & \{1, 1, 1, 1, 1, 1, 1, 1, 1, 1, 1, 1, 1, 1\} & $0.98\pm0.94$ & $1.31\pm2.29$ & $0.71\pm0.66$ & $5.21\pm4.01$ & $\mathbf{1.56\pm1.40}$ & $10.37\pm5.36$ & $10.88\pm9.15$ & $2.05\pm2.94$ & $4.43\pm5.94$\\
            8 & 1.1 & 1.1 & \{0, 0, 0, 1, 1, 1, 1, 1, 1, 1, 1, 0, 0, 0\} & $1.12\pm1.71$ & $0.98\pm0.65$ & $2.98 \pm 4.50$ & $4.55 \pm 0.87$ & $3.14 \pm 1.63$ & $10.44\pm4.99$ & $13.89\pm9.19$ & $2.40 \pm 2.79$ & $5.30 \pm 6.25$\\
            9 & 1.1 & 1.1 & \{0, 0, 0, 0, 0, 0, 0, 0, 0, 0, 0, 0, 0, 0\} & $0.76 \pm 0.94$ & $1.09 \pm 1.29$ & $3.41\pm2.78$ & $3.46 \pm 3.84$ & $5.81 \pm 4.43$ & $7.83 \pm 4.74$ & $\mathbf{10.65\pm4.36}$ & $2.18 \pm 2.72$ & $4.71\pm4.75$\\

            \bottomrule 
        \end{NiceTabular}
    }
    \label{table:sim2real_results}
\end{table*}

\subsection{Tool Optimisation via BOHB} 
\label{ssec:bohb}

We ran a BOHB-driven search, as described in Section~\ref{sec:methodology}, over the morphological design space $\Xi$. 
We used the BOHB implementation from SMAC3~\cite{smac3}, with a random forest as the surrogate model and expected improvement (EI) as the acquisition function. 
We set the successive halving proportion to $\eta=3$, with a minimum budget $b_{\mathrm{min}}=330$ episodes and a maximum budget $b_{\mathrm{max}}=3000$ episodes. 
We provide the configurations evaluated in Section~\ref{ssec:grid_search} as warm-start points for the optimiser. 
When a configuration is promoted to a higher budget, training resumes from its checkpoint rather than re-initialising. 
All other hyperparameters follow Section~\ref{ssec:grid_search}.
We evaluate $n=2$ random seeds per configuration at the budget $b$ prescribed by successive halving. 
Accounting for each seed independently, the total $200,000$-episode optimisation budget required approximately $7.85$ days of compute across four parallel workstations.

\begin{figure}[htbp]
    \centering
    \renewcommand{\arraystretch}{1.2}
    \setlength{\tabcolsep}{2pt}
    \begin{tabular}{@{} c c c @{}}

        \includegraphics[width=0.27\linewidth]{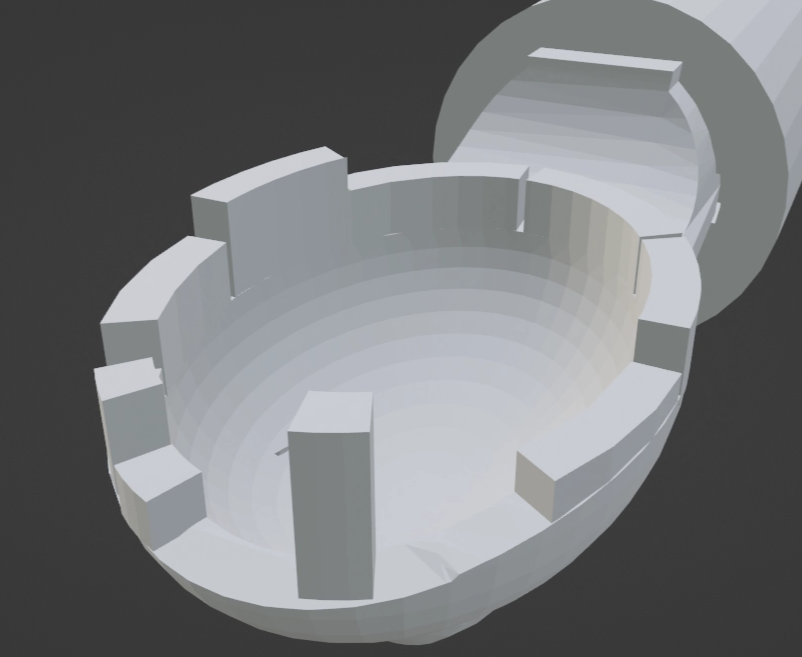} &
        \includegraphics[width=0.27\linewidth]{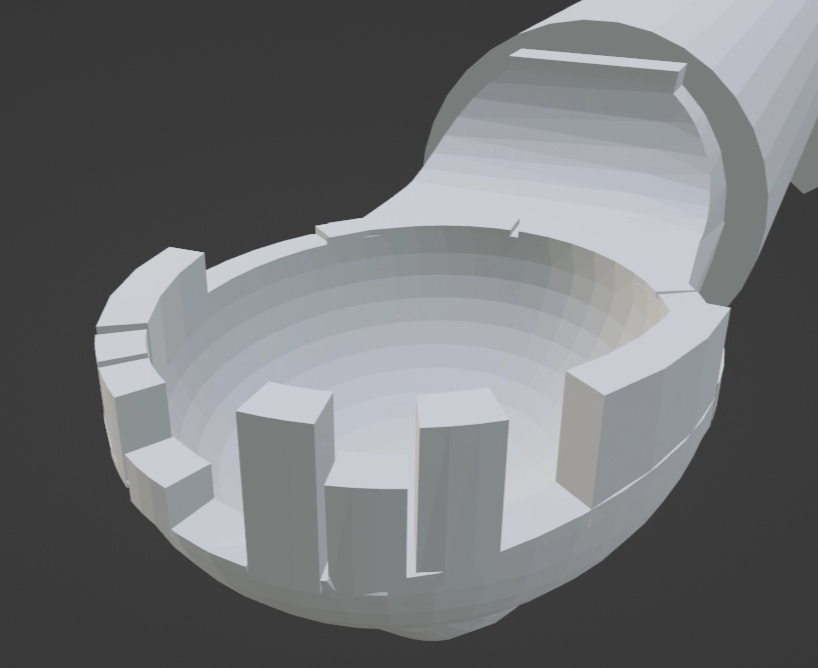} &
        \includegraphics[width=0.27\linewidth]{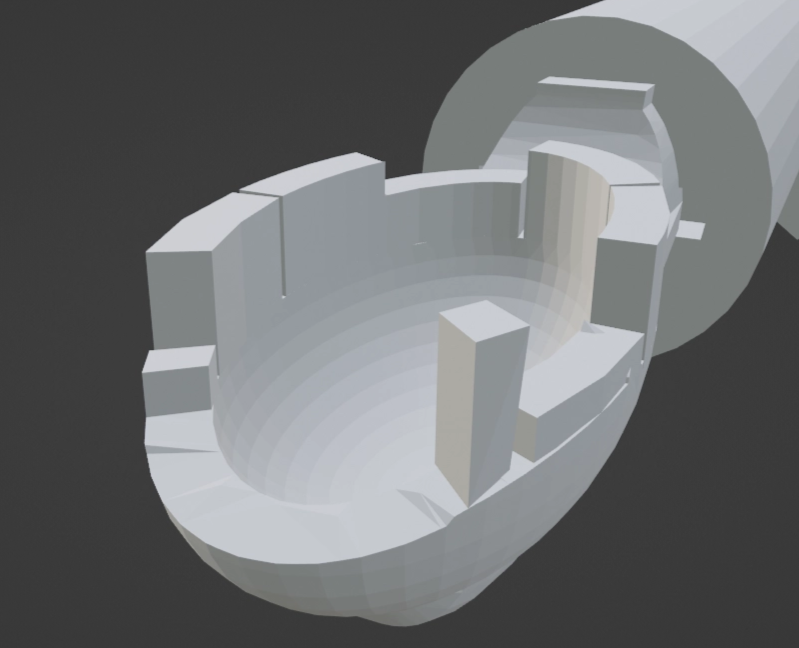} \\
        \includegraphics[width=0.27\linewidth]{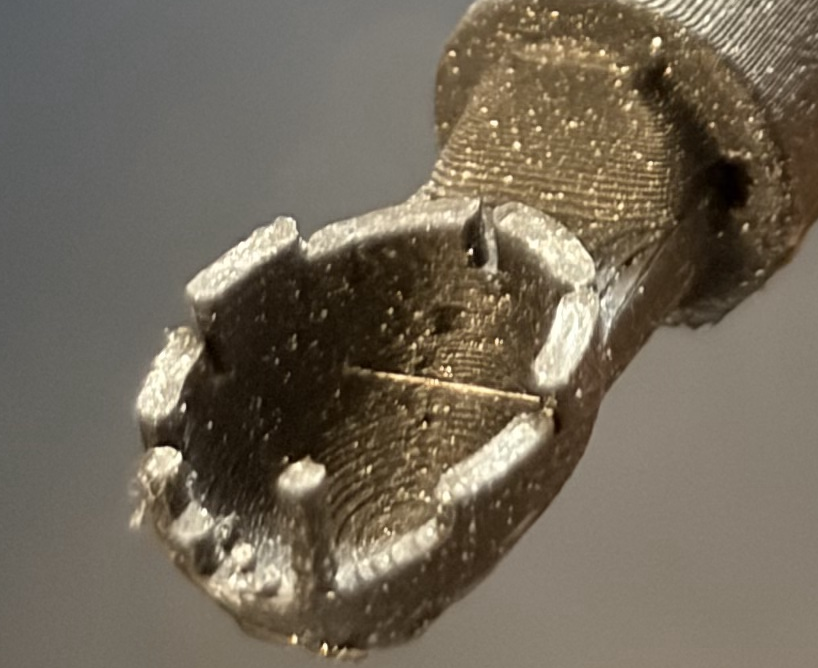} &
        \includegraphics[width=0.27\linewidth]{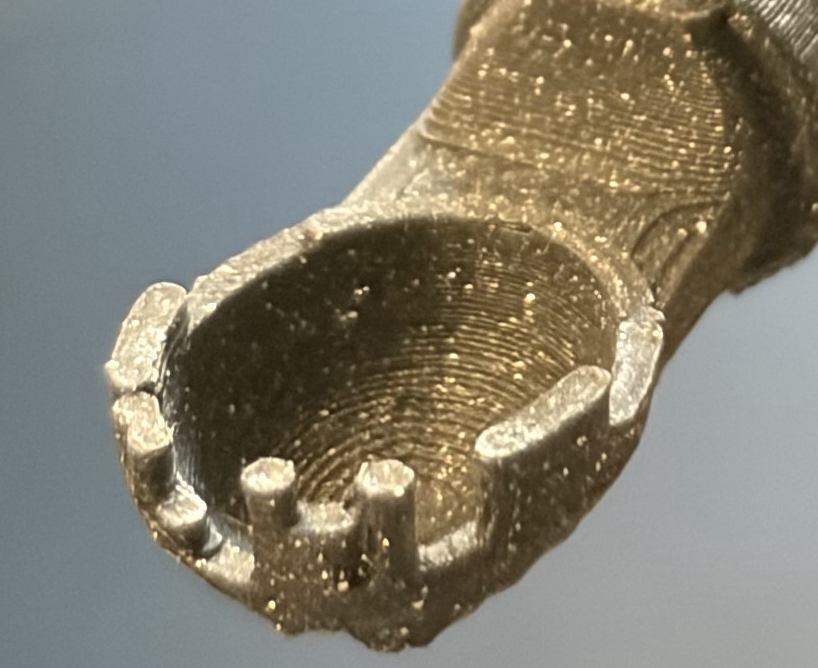} &
        \includegraphics[width=0.27\linewidth]{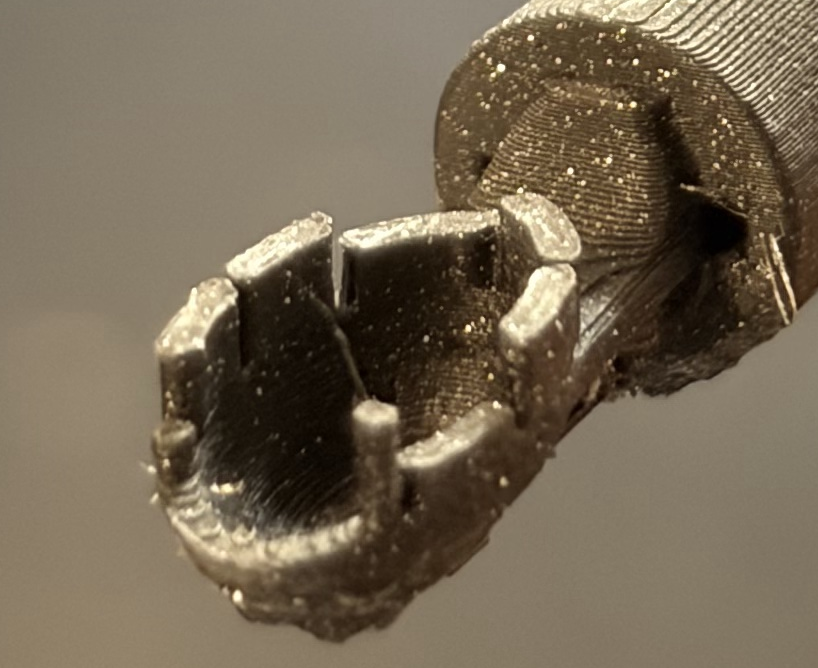} \\
        \textbf{a} & \textbf{b} & \textbf{c} \\
    \end{tabular}

    \vspace{10pt}
    \renewcommand{\arraystretch}{1.2}
    \resizebox{\columnwidth}{!}{
        \begin{tabular}{c c c l}
            \hline
            \textbf{Morph.} & \textbf{Width} & \textbf{Depth} & \textbf{Spike Configuration} \\
            \hline
            \textbf{a} & 0.92 & 1.14 & \texttt{\{0,0.33,0.66,0.33,0.66,1,0.66,0.66,0.66,0.33,0,0,1,0\}} \\
            \textbf{b} & 1.08 & 1.28 & \texttt{\{0,0.33,0.66,0.66,0.66,0,0.33,0.33,0.33,1,0,1,0.66,1\}} \\
            \textbf{c} & 0.70 & 1.13 & \texttt{\{0,0,0,0.33,1,1,0.66,1,1,0.33,0,0,0,0\}} \\
            \hline
        \end{tabular}
    }

    \caption{Top three morphologies discovered by the BOHB optimiser. The simulated models are shown in the top row, with their corresponding physical 3D-printed prototypes below. The table details the specific width, depth, and spike configurations for each design.}
    \label{fig:2x3_grid}
\end{figure}
\begin{table}[h]
    \caption{
    Simulation and real-world deployment comparison of the best 3 tool configurations found by BOHB, reported from the high-fidelity simulation environment and for IID powders, averaged over 10 real-world runs for $w_{target}=15$\SI{}{mg}.}
    \centering
    \label{table:plain_bohb_s2r-results}
    \resizebox{0.99\linewidth}{!}{
        \begin{NiceTabular}{l *{8}{c}}[hlines, cell-space-limits=3pt]
            \hline 
            \Block{2-1}{Configuration} & \multicolumn{4}{c}{\textbf{Powder Weighing Error (mg)}} & \Block{2-1}{Simulation Error}  & \Block{2-1}{Real-world Error} \\
             & Sand & Sugar & Salt & Semolina & & \\
             \hline
            \midrule 
            Config \textbf{a} & $0.80\pm0.45$ & $0.96\pm0.89$ & $1.03\pm0.89$ & $2.58\pm1.58$ & $0.97\pm0.31$ & $1.31\pm1.17$\\
            Config \textbf{b} & $\mathbf{0.55 \pm 0.24}$ & $\mathbf{0.71 \pm 0.35}$ & $\mathbf{0.65\pm0.73}$ & $1.85\pm1.43$ & $0.94\pm 0.09$&$\mathbf{0.91\pm0.93}$ \\
            Config \textbf{c} & $0.76 \pm0.58$ & $1.0\pm0.9$ & $1.18\pm1.44$ & $\mathbf{1.52\pm1.36}$ & $\mathbf{0.85\pm0.17}$ & $1.10\pm1.10$\\
            \bottomrule 
        \end{NiceTabular}
    }
\end{table}

We selected the top three morphologies identified by the BOHB optimiser, ranked by their weighing error in the high-fidelity environment at the maximum budget $b_{\mathrm{max}}$. 
For these three designs, we train two additional random seeds, bringing the total to four per configuration. 
We report the simulated error as the average across four seeds and deployed the best-performing policy from simulation for each morphology to the physical robot. 
Fig.~\ref{fig:2x3_grid} illustrates the three designs in simulation alongside their 3D-printed counterparts, with their morphological configurations. 
Table~\ref{table:plain_bohb_s2r-results} compares the simulated error against the average real-world error. 
We report the real-world error only for in-distribution materials, as the findings in Section~\ref{ssec:grid_search} demonstrated that zero-shot transfer for OOD, highly cohesive materials that are poorly modelled in our simulation framework is unreliable. 

All three configurations are predicted by the simulator to outperform the best tool of Section~\ref{ssec:grid_search}. 
However only Config. \textbf{b} and Config. \textbf{c} achieve lower real-world errors. 
Config. \textbf{b} demonstrates an overall improvement of \SI{0.26}{mg} across the in-distribution powders, exhibiting better performance on sand, salt, and semolina, while Config. \textbf{c} achieves the lowest absolute error on semolina. 
Although the optimiser predicted Config. \textbf{c} to be the best design overall, Config. \textbf{b} outperformed it in physical trials and exceeded its own simulated prediction. 
The sim-to-real prediction gaps for these configurations nonetheless remain small (\SI{0.25}{mg} for Config. \textbf{c} and \SI{0.03}{mg} for Config. \textbf{b}). 
We attribute this rank inversion to the reality gap between the simulator and physical setup, which includes unmodelled dynamics and morphological variations introduced by 3D printing.

\subsection{Accelerated Co-design via Morphological Similarity}
\label{ssec:bohb_similarity}

\begin{figure}[h!]
    \centering
    \includegraphics[width=\linewidth]{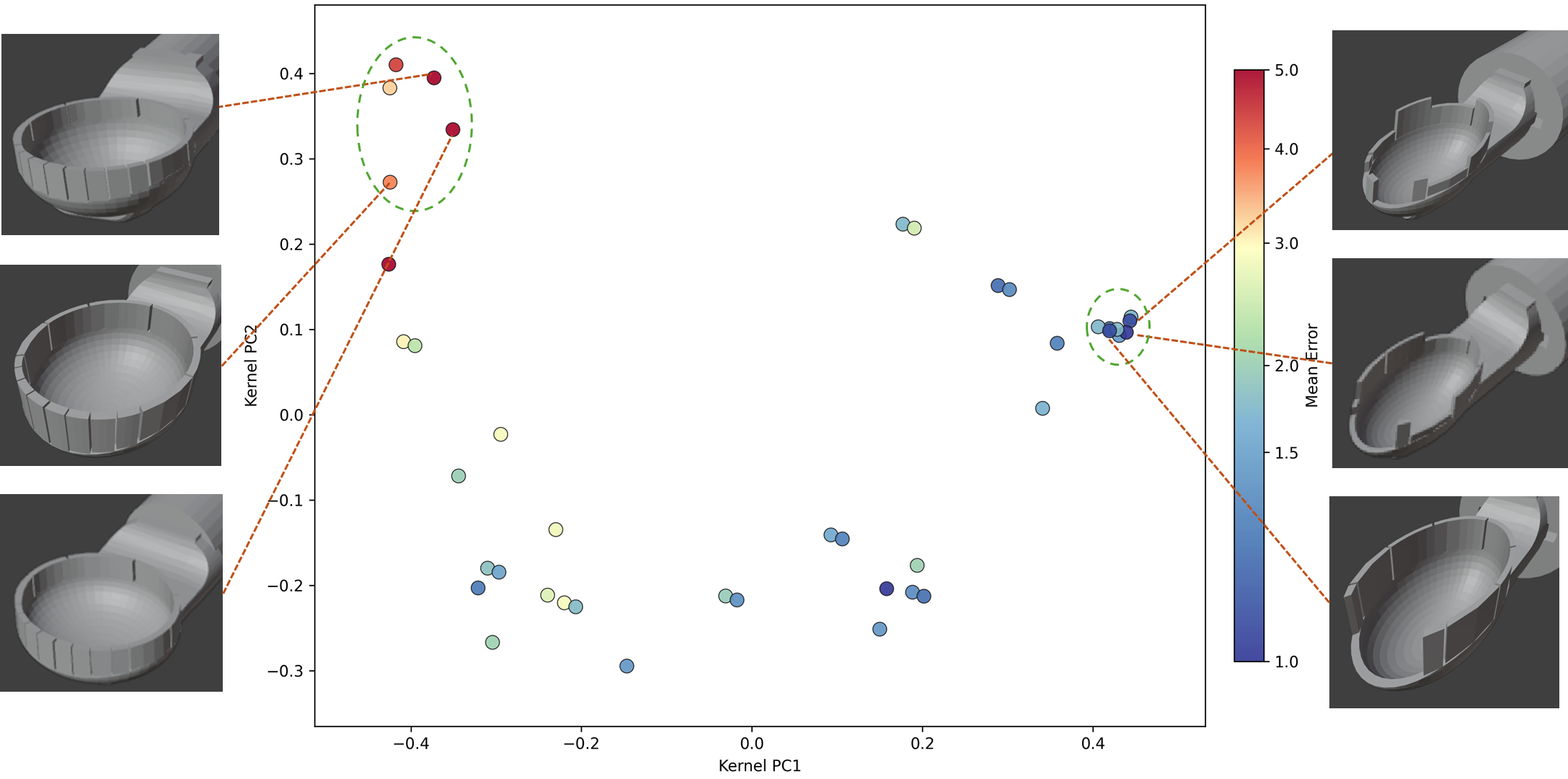}
    \caption{Each point is two-dimensional representation of a spoon configuration. Spoon designs that are located closer together in the 2D space have higher morphological similarity. The colour of the points represents their task error in simulation.}
    \label{fig:similarity_mapping}
\end{figure}

We evaluated the similarity-based strategy introduced in Section~\ref{ssec: similarity}, to determine whether it accelerates the optimisation process without degrading the real-world performance of the discovered tools. 
To compute the spike similarity score $S_{\mathrm{spikes}}$, we defined the spike arc ratios as $c_i \in \{1/32, 1/8\}$. 
The overall similarity $S(\xi_1, \xi_2)$ was then computed using Equation~\ref{eq:sim_total}, with the coefficients $\alpha$, $\beta$, and $\gamma$ tuned to $0.5$, $0.3$, and $0.2$, respectively. 
To examine whether the metric captures task-relevant structure, we aggregated the morphologies from the initial grid search (Section~\ref{ssec:grid_search}) and the maximum budget evaluations from the BOHB optimisation (Section~\ref{ssec:bohb}), and projected them into two dimensions using kernel principal component analysis (KPCA)~\cite{scholkopf1998nonlinear}, with $S(\xi_1,\xi_2)$ supplied as a precomputed kernel instead of a distance in the raw parameter space. 
Fig.~\ref{fig:similarity_mapping} illustrates the resulting distribution of the dimensionality-reduced morphologies, mapped alongside their corresponding simulated weighing errors. 
Morphologies that cluster in the embedding, which appear structurally similar upon visual inspection, exhibit comparable task performance, indicating that the proposed similarity is correlated with dispensing behaviour. 
This supports using this metric to warm-start and accelerate policy training.  

\begin{figure}[h!]
    \centering
    \includegraphics[width=0.85\linewidth]{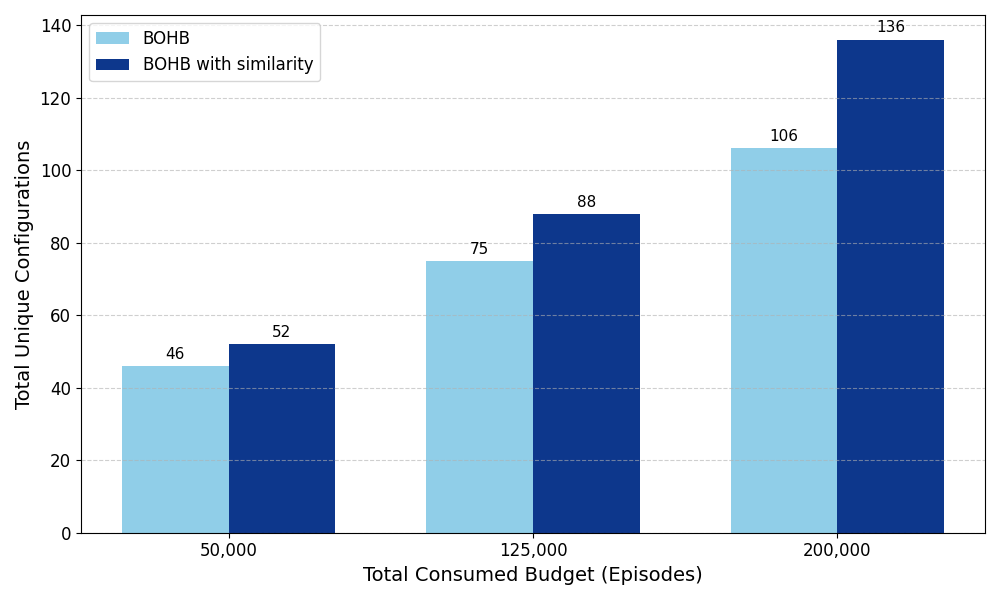}
    \caption{BOHB performance with and without morphology similarity filter. The plot shows total unique sampled configurations at $50000$, $125000$ and $200000$ total episodes used.}
    \label{fig:bohb comparisons}
\end{figure}

We repeated the search from Section~\ref{ssec:bohb} using identical hyperparameters, with the similarity-based logic from Algorithm~\ref{alg:bohb_spoon}. 
The similarity threshold $\sigma$ and warmup fraction $\phi$ were set to $0.85$ and $0.33$, respectively. 
Figure~\ref{fig:bohb comparisons} compares unique configurations explored within the fixed budget of $200,000$ episodes by the BOHB (Section~\ref{ssec:bohb}) and its similarity-augmented variant. 
Both methods exhibit similar initial exploration rates, but the similarity-augmented BOHB accelerates as the number of sampled configurations increase. 
Ultimately, the similarity approach explores $136$ unique configurations, compared to $106$ for the standard baseline. 
As the cached policy pool grows, the warm-start hit-rate increases, reducing the average training cost per configuration.

Following the evaluation protocol of Section~\ref{ssec:bohb}, we select the top three configurations, train two additional random seeds each and deploy the best-performing policy on the physical robot. 
Table~\ref{table:bohb-similarity-configs} reports the sim-to-real transfer with the predicted simulation errors. 
All three morphologies outperformed the grid-search tool of Section~\ref{ssec:grid_search} in simulated and physical experiments, with predicted errors comparable to the standard BOHB run.
Notably, Config. \textbf{B} is morphologically identical to Config. \textbf{b} from Section~\ref{ssec:bohb}. 
Upon sim-to-real transfer, Config. \textbf{B} now shows a minor improvement of \SI{0.10}{mg} weighing error, emerging as the best-performing physical morphology in this experiment as well. 
While this variance in physical performance is likely attributable to policy seeding, the convergence on the same geometry under both settings indicates that similarity-based warm-starting accelerates morphological search without degrading the quality of the final tool.

\begin{table}[h]
    \caption{Simulation and real-world deployment comparison of the best 3 tool configurations found by BOHB with similarity, reported from the high-fidelity simulation environment and for IID powders, averaged over 10 real-world runs for $w_{target}=15$\SI{}{mg}.}
    \centering
    \label{table:bohb-similarity-configs}
    \resizebox{0.99\linewidth}{!}{
        \begin{NiceTabular}{l *{8}{c}}[hlines, cell-space-limits=3pt]
            \hline 
            \Block{2-1}{Configuration} & \multicolumn{4}{c}{\textbf{Powder Weighing Error (mg)}} & \Block{2-1}{Simulation Error}  & \Block{2-1}{Real-world Error } \\
             & Sand & Sugar & Salt & Semolina & & \\
             \hline
            \midrule 
            Config \textbf{A} & $0.65\pm0.54$ & $\mathbf{0.47\pm0.24}$ & $0.70\pm0.36$ & $1.82\pm1.25$ & $0.95\pm0.05$ & $0.91\pm0.87$\\
            Config \textbf{B} & $0.62 \pm 0.55$ & $0.56 \pm 0.23$ & $\mathbf{0.41\pm0.18}$ & $\mathbf{1.66\pm0.92}$ & $0.92\pm 0.08$&$\mathbf{0.81\pm0.73}$ \\
            Config \textbf{C} & $\mathbf{0.59 \pm0.27}$ & $0.73\pm0.50$ & $0.91\pm1.33$ & $1.97\pm0.91$ & $\mathbf{0.86\pm0.22}$ & $1.05\pm0.98$ \\
            \bottomrule 
        \end{NiceTabular}
    }
    
    \vspace{4mm}
    
    \resizebox{\linewidth}{!}{
        \begin{tabular}{c c c l}
                \hline
                Config. & Width & Depth & Spike Configuration\\
                \hline
                \textbf{A} & 0.80 & 1.45 & \texttt{\{0,0.66,0.66,0.0,1,1,0,0.33,1,0.66,0.66,0.66,0.66,0.33\}} \\
                \textbf{B} & 1.08 & 1.28 & \texttt{\{0,0.33,0.66,0.66,0.66,0,0.33,0.33,0.33,1,0,1,0.66,1\}} \\
                \textbf{C} & 1.02 & 1.32 & \texttt{\{0,0.66,0,0.66,1,0,0.33,0.33,0,0.33,0.33,0.66,0,1\}} \\
                \hline
            \end{tabular}
    }
\end{table}
\subsection{Discussion}
Our framework identifies tools that outperform human-centric standard ones and manually-designed baselines. 
While our high-fidelity simulation predicts real-world performance for IID materials ($28^\circ$ to $41^\circ$ AoR), we deployed our best spoon (Config. \textbf{B}) on OOD materials to evaluate generalisability (Table~\ref{table:ood_evaluation}). 
As material dynamics drift from the training distribution, the control policy becomes less reliable as the sim-to-real gap widens. 
Despite this degradation on OOD powders, Config. \textbf{B} achieves lower overall errors than the baselines across the combined set.
While our co-designed tool is specialised for its target distribution of powder dynamics, it remains a more effective all-purpose tool for robotic chemists than standard human tools.
\begin{table}[htbp]
    \centering
    \caption{Zero-shot transfer results on OOD materials with their respective AoR, averaged over 10 runs for $w_{target}=15$\SI{}{mg}.}
    \label{table:ood_evaluation}

\resizebox{0.99\linewidth}{!}{
    \begin{tabular}{l*{5}{c}}
        \toprule
    
        Morphology & IID Error & Sodium Bicarbonate ($43^\circ )$ & Pectin ($46^\circ$) & Flour ($51^\circ$) & Overall Real-world Error\\ \midrule
        Standard Tool & $1.93 \pm 2.16$ & $4.45 \pm 5.57$ & $3.58 \pm 3.64$ & $12.91 \pm 13.58$ & $4.09 \pm6.82$ \\
        Config. 2 & $1.17\pm1.13$ & $3.62\pm3.45$ & $8.21 \pm 4.04$ & $11.41\pm8.72$ & $4.04\pm 5.43$
        \\
        Config. B & $\mathbf{0.81 \pm 0.73}$ & $\mathbf{1.98 \pm 1.53}$ & $\mathbf{2.71 \pm 2.43}$ & $\mathbf{7.69 \pm 4.01}$ & $\mathbf{2.23 \pm 3.00}$ \\ \bottomrule
    \end{tabular}
}
\end{table}
\section{Conclusion}
We presented a co-design framework for autonomous powder weighing that jointly optimises a tool's morphology and its reinforcement learning control policy, using a geometric similarity metric to warm-start the search from structurally related candidates. 
Real-world experiments show the co-designed tools outperform both a standard tool and grid-search baselines across materials.
The main limitation is that our simulation cannot accurately reproduce cohesive powders; future work will develop higher-fidelity granular simulation that remains tractable for search.
Ultimately, extending joint morphology–control optimisation to other contact-rich laboratory tasks will improve robot–material manipulation in autonomous scientific discovery.


\vspace{-0.5em}


\bibliographystyle{IEEEtran}
\bibliography{bibliography}

\end{document}